\documentclass[
]{ceurart}

\usepackage{listings}
\usepackage{caption}  
\begin{document}

\copyrightyear{2025}
\copyrightclause{Copyright for this paper by its authors.
  Use permitted under Creative Commons License Attribution 4.0
  International (CC BY 4.0).}

\conference{CLEF 2025 Working Notes, 9 -- 12 September 2025, Madrid, Spain}

\title{\textsc{MaLei} at MultiClinSUM: Summarisation of Clinical Documents using Perspective-Aware Iterative Self-Prompting with LLMs}

\title[mode=sub]{Notebook for the <MultiClinSUM> Lab at CLEF 2025}


\author[1,2]{Libo Ren}[%
orcid=0009-0005-9182-9424,
email=renlibo994@gmail.com,
url=https://github.com/Libo-Ren,
]
\address[1]{University of Manchester, UK}
\address[2]{Modul University Vienna, Austria}

\author[3]{Yee Man Ng}[%
orcid=0009-0008-9659-6510, 
email=y.m.ng@liacs.leidenuniv.nl,
url=https://www.universiteitleiden.nl/en/staffmembers/yee-man-ng
]
\address[3]{Leiden Institute of Advanced Computer Science (LIACS), Leiden University, The Netherlands}

\author[1,3,4]{Lifeng Han}[%
orcid=0000-0002-3221-2185,
email=l.han@lumc.nl,
url=https://www.universiteitleiden.nl/en/staffmembers/lifeng-han,
]
\cormark[1]
\address[4]{Leiden University Medical Center, The Netherlands}


\cortext[1]{Corresponding author: LH.}

\begin{abstract}
Efficient communication between patients and clinicians plays an important role in shared decision-making. However, clinical reports are often lengthy and filled with clinical jargon, making it difficult for domain experts to identify important aspects in the document efficiently. 
This paper presents the methodology we applied in the MultiClinSUM shared task for summarising clinical case documents.
We used an Iterative Self-Prompting technique on large language models (LLMs) by asking LLMs to generate task-specific prompts and refine them via example-based few-shot learning.
Furthermore, we used lexical and embedding space metrics, ROUGE and BERT-score, to guide the model fine-tuning with epochs.
Our submission using perspective-aware ISP on GPT-4 and GPT-4o achieved ROUGE scores (46.53, 24.68, 30.77) and BERTscores (87.84, 83.25, 85.46) for (P, R, F1) from the official evaluation on 3,396 
clinical case reports from various specialties extracted from open journals.
The high BERTscore indicates that the model produced semantically equivalent output summaries compared to the references, even though the overlap at the exact lexicon level is lower, as reflected in the lower ROUGE scores.
This work sheds some light on how perspective-aware ISP (PA-ISP) can be deployed for clinical report summarisation and support better communication between patients and clinicians.
  
\end{abstract}


\maketitle

\section{Introduction}
Efficient and effective communications between patients and healthcare professionals play an important role in patient care \cite{stiggelbout2015shared,stiggelbout2023metro}. However, 
healthcare providers frequently 
have to read many clinical documents in a short time frame to understand the current patients. This is challenging, as clinical documents of patients include rich information on patients’ problems, diagnoses, treatments, progressions, and side effects. Similarly, patients often do not have the clinical expertise to fully understand the lengthy clinical documents about their health issues. 
A concise and accurate summarisation of clinical documents will save the time of healthcare professionals to understand the problem at hand, as well as help patients to understand their health conditions better and earlier. 
We attended the multilingual clinical documents summarisation shared task (MultiClinSUM) to explore large language models (LLMs) for the use of this challenge.

The MultiClinSUM Track is organised by the Barcelona Supercomputing Center’s NLP for Biomedical Information Analysis group and promoted by Spanish and European projects such as DataTools4Heart, AI4HF, and BARITONE \url{https://temu.bsc.es/multiclinsum/}.  

To utilise the current state-of-the-art development from natural language processing (NLP), we investigated Iterative Self-Prompting (\textbf{ISP}) \cite{romero2025manchester} in GPT-4 and ChatGPT for automatic summary of clinical documents.
In this methodology, we ask the LLMs to generate prompts themselves to approach this task by detailed instruction and example-based learning; the generated automatic summary is returned to the LLMs with example references to ask the LLM to refine the prompts for this specific task.
The ISP technique has proved to be very useful and efficient in leveraging LLMs to generate a clear summary that includes patients' symptoms, diagnoses, treatments, and outcomes/follow-ups, in a previous shared task for healthcare answer summarization \cite{romero2025manchester}.
We use the datasets provided by the shared task organizers, which contain clinical case reports and summaries that are written by healthcare providers and in various languages, including English, Spanish, French, and Portuguese. 
The \textbf{English} test set under investigation contains 3,396 clinical documents.
We perform both quantitative and qualitative evaluations, through human and automatic metrics, including ROUGE and BERTscore. ROUGE is a lexical overlapping metric, and BERTscore is an embedding space semantic similarity metric.

Quantitative evaluation scores show that while our system output has a ROUGE F1 score of 0.31 compared to the reference on exact lexicon matches, the semantic BERTscore shows 0.85 F1, which indicates high quality of semantic meaning preservation. This sheds good light on the potential usage of LLMs for the summarisation of clinical documents using the ISP technique.
The qualitative analysis confirms that the generated summaries tend to cover the key clinical aspects and contain logical paraphrasing. 
We also carried out error analysis to see in what ways LLMs produce undesirable results, such as the considerable length of the summaries.
Another interesting finding from this shared task is that LLMs tend to generate longer text to comment on missing data when the clinical document is too short, such as ``The case report does not provide specific details on the outcome or follow-up. Typically, such a patient would require close monitoring and treatment adjustments based on laboratory and clinical responses.'' This also provides some insight into the clarity of current clinical documents/reports.


\section{Background and Related Work}
\subsection{Clinical NLP}
Clinical NLP has drawn attention from both NLP and healthcare researchers in recent years due to the development and effectiveness of modern NLP models and the eagerness to test such AI models in healthcare domains. For instance, there is the ClinicalNLP workshop series from 2016 (6th edition in 2024) \cite{clinicalnlp-2024-clinical}. 
A related track of ClinicalNLP WS is the Biomedical NLP international workshop events (BioNLP) from 2004 \cite{ws-2004-international,bionlp-2024-biomedical}.

The corresponding tasks have included clinical text translation \cite{han2024neural}, biomedical abstract simplification \cite{ling2024malei,li2024investigating}, clinical events recognition \cite{belkadi2023exploring}, temporal relation extraction \cite{tu2023extraction,cui2023medtem2}, entity linking/normalisation to SNOMED CT and  British National Formulary (BNF) codes \cite{romero2025medication}, 
synthetic data generation/augmentation \cite{ren2025synthetic4health,ren2025beyond}, patient sensitive information de-identification \cite{paul2024deidclinic},
and healthcare answer summarisation \cite{romero2025manchester}, etc. 
These works have explored methodologies from different paradigms, such as fine-tuning encoder-based models, training encoder-decoder models, and prompting decoder-only models using different techniques.

\subsection{Healthcare Data Summarisation}
For the clinical documents summarisation task, 
the most relevant work includes the 
shared task on Perspective-Aware Healthcare Answer Summarisation (PerAnsSumm 2025) \cite{agarwal-etal-2025-overview}. This shared task included the summarisation of online forum healthcare answers while considering the different perspectives, i.e., types of information such as `Cause' or `Suggestion', within an answer. 
In this shared task, we used Iterative Self-Prompting (ISP) with Claude and o1 for Perspective-aware Healthcare Answer Summarisation, as described in \cite{romero2025manchester,agarwal-etal-2025-overview}.
Similar to clinical documents, online forum responses vary greatly in length.
The key difference is that clinical documents often contain clinician-specific abbreviations and jargon, which pose challenges for NLP models to interpret. In contrast, online forum data typically includes social media-style writing, such as frequent spelling mistakes and grammatical errors.
In this work, we build upon our experience from the PerAnsSumm shared task to design perspective-aware summaries for clinical documents.

\section{Methodology}
\label{sec: Methodology}
\subsection{Overview of the Prompting Framework}
There are mainly three clinical datasets involved in this project, as listed below:
\begin{itemize}
    \item \textit{multiclinsum\_gs\_train\_en: } contains 592 gold-standard samples, which are manually annotated and consist of a full-text and summary pair.
    \item \textit{multiclinsum\_large-scale\_train\_en: } contains 25,902 full-text and summary pairs. Their quality is slightly lower than that of the 592 gold samples, but they are still useful for data augmentation.
    \item \textit{multiclinsum\_test\_en: } the English test set, which includes 3,396 full-text cases but without any summaries.
\end{itemize}

\begin{figure}
  \centering
  \includegraphics[width=\linewidth]{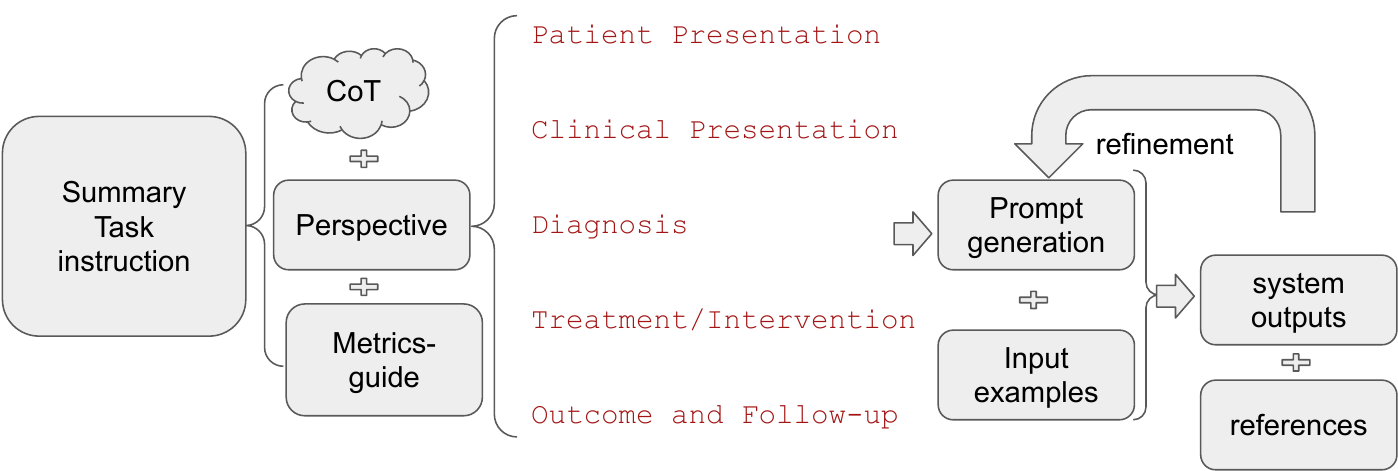}
  \caption{Perspective-aware Iterative Self-Prompting (PA-ISP) Illustration Diagrams from the \textsc{MaLei} team.}
  \label{fig:ISP}
\end{figure}

We mainly adopt the Iterative Self-Prompting (ISP) strategy in this task, with some task-specific features. As shown in Figure \ref{fig:ISP}, we construct the meta-prompt based on the combination of Chain-of-Thought (CoT) instructions, clinical perspectives, and metric-based guidance. The meta-prompt is provided to the LLM along with a few few-shot examples at the beginning.

Based on this meta-prompt and the examples, the LLM is instructed to generate a new task-specific prompt that guides the clinical summarisation process more effectively. Using this synthetic prompt (prompt\_v1), we input it together with clinical full-texts from a portion of the golden training set into the model to generate corresponding summaries. These synthetic summaries are then compared with the ground-truth summaries from the golden data, and evaluation scores—as well as reflective feedback and suggestion advice—are produced accordingly. This feedback serves as a reference for further prompt refinement, allowing us to iteratively update the prompt to obtain prompt\_v2, \_v3, and so on.

This prompt updating process is repeated until no obvious performance improvement is observed. Once the improvement plateaued, we planned to augment the structure using additional spans extracted from the remaining gold data and to apply a retrieval-based (RAG) technique using the 25,902 non-golden training samples. After all experiments, the best-performing prompt version is used to generate the final clinical summaries on the test set.

Unfortunately, due to time constraints, we were only able to complete the steps before the structure augmentation phase  (as detailed in Section \ref{sec: Full-data Structure-aware Prompt Enhancement}). As a result, we selected the best-performing prompt at that point and used it directly for inference on the test set. The remaining experimental designs can be explored in future work.

\subsection{Prompt Initialisation and Few-shot Setup}
\label{sec:Prompt Initialisation and Few-shot Setup}
As shown in Figure \ref{fig:ISP}, we first construct 
an initial instruction that prompts the LLMs about the summary task description with the chain of thoughts (CoTs) on how it shall think, for example:
\begin{itemize}
    \item What common structure or patterns do you observe in the examples?
    \item What information is emphasised?
    \item How can a language model be guided to produce similar quality outputs?
    \item What errors should be avoided?
\end{itemize}
These CoTs are combined with perspective-based, i.e., multifaceted, structural guidance and metric-based feedback to inform the LLM’s generation process.

The Perspectives we designed include:
\begin{enumerate}
    \item Patient Presentation: age, sex, relevant history.
    \item Clinical Presentation: key symptoms and signs.
    \item Diagnosis: relevant investigations, tests, conclusions.
    \item Treatment/Intervention: medications, surgeries, therapies.
    \item Outcome and Follow-up: results of treatment, current status.
\end{enumerate}

The metrics we used are ROUGE-L (for lexical overlap) and BERTScore (for embedding semantic fidelity). The instructions, structural perspectives, evaluation metrics, and three representative examples collectively form the meta-prompt, which is then used to generate the initial prompt. An example of the meta-prompt is provided in Appendix \ref{sec:prompt-example}.

\subsection{Prompt Iterative Update and Refinement}
We selected a small batch of 50 full-text and summary pairs to train the prompt-interactive model. Note that data points indexed from 4 to 53 were used at this stage, as the first three samples had already been used for the initial prompt generation as few-shot examples. 

In each epoch, we not only generated summaries based on the full texts using the initial prompt but also compared the synthetic summaries with the gold-standard annotations, asking GPT-4o to provide reflections and revision suggestions. For each generated summary, we computed both ROUGE-L and BERTScore, and requested reflections and revision suggestions from the model behind their performance. We found that BERTScore remained relatively stable (consistently above 0.85), while ROUGE-L scores fluctuated significantly, ranging from 0.12 to 0.52. Therefore, our optimisation efforts focused on improving ROUGE-L.

To iteratively update the prompt while balancing performance and computational cost, we selected a small subset of 15 summaries with the lowest ROUGE-L scores and included their corresponding evaluation feedback. These were used as new few-shot examples to guide the prompt refinement. The LLM was then instructed to revise the prompt by integrating the previous version along with the reflection and suggestion content of these samples. The prompt used in this process is shown in the Appendix \ref{sec: Instruction for Prompt Refinement} Figure \ref{fig:Instruction for Prompt Refinement}.

We conducted five epochs of this process. For the initial version (prompt\_v1), one summary was found to be invalid for evaluation scoring, with an overall BERTScore of 0.86 and a ROUGE-L of 0.30 among the remaining 49 full-text and summary pairs. Interestingly, after the prompt updates, although the evaluation scores did not improve significantly, the invalid case was resolved. In other words, the BERTScore continued to fluctuate around 0.86, and the ROUGE-L around 0.30, across all 50 full-text and summary pairs. As a result, we adopted prompt\_v2 as the best-performing version—the first in which invalid predictions were eliminated.

\subsection{Full-data Structure-aware Prompt Enhancement}
\label{sec: Full-data Structure-aware Prompt Enhancement}
To further improve structural consistency, the remaining 539 gold summaries were intended for extracting common phrases and analysing section-wise linguistic patterns. For example, we aimed to identify common clinical spans and examine the language style of the gold-standard summaries
, e.g., phrases like \textit{“The patient presented with…”} or \textit{“Treatments include…”}. We also considered measuring the average length of each paragraph and investigating whether consistent structural patterns could be observed and used to refine the prompt.

Based on these insights, the native instructions for the five clinical perspectives could be further specified. In addition, regular expressions or phrase-matching methods could be employed to capture high-frequency sentence structures or templates. Sentence patterns that are often overlooked could also be statistically analysed. Combining these steps may contribute to improving evaluation scores, particularly the ROUGE-L score.

Despite several iterations of prompt revision based on reflective feedback, the model outputs did not demonstrate any improvements in ROUGE-L, indicating that more sophisticated strategies beyond self-iterative prompt refinement may be required. Owing to time constraints, further exploration in this area is reserved for future work.

\subsection{Similar Case Retrieval-based Few-shot Augmentation}
\label{sec: Similar Case Retrieval-based Few-shot Augmentation}
In addition to the gold-standard set, the dataset also includes 51,804 extended clinical cases, each consisting of a full-text input and its corresponding summary. To enhance test-time generation, we retrieve cases whose input texts are semantically similar to the current full-test input using sentence embeddings. SentenceBERT, combined with cosine similarity, is applied at this stage.

The summaries from the top retrieved cases are inserted before the test input as few-shot demonstrations, following the same format as the manually selected gold examples at stage \ref{sec:Prompt Initialisation and Few-shot Setup}. Typically, the top 3 cases are selected to balance diversity with prompt length constraints. The retrieved summaries are not included in the evaluation and serve solely as auxiliary input for generation guidance.

\subsection{Testset Inference}
Once the optimal prompt was selected, the final version was used for testset summarization. The generation process follows the same setup described in Section \ref{sec:Prompt Initialisation and Few-shot Setup}. The main difference is that we expect the generated summary to be shorter than the original full text. Therefore, we compared the character lengths of the synthetic summaries and their corresponding full texts and identified cases where the summary was unexpectedly longer.

For these cases, we asked the language model to regenerate the summary up to five times. Some outputs were successfully shortened, while others remained longer than the input. In such cases, we directly replaced the generated summary with the original full text, assuming that the original text was already sufficiently concise.

It should be noted that although the full experimental pipeline was initially designed, we did not proceed with the Full-data Structure-aware Prompt Enhancement and Similar Case Retrieval-based Few-shot Augmentation (as detailed in Section \ref{sec: Full-data Structure-aware Prompt Enhancement} and Section \ref{sec: Similar Case Retrieval-based Few-shot Augmentation}) due to time constraints. These steps are planned for future work to further explore their potential for improving performance.

\section{Experimental Work and Submission to MultiClinSUM}
\subsection{Development of the ISP-GPT-4/o model}
\begin{figure}
  \centering
  \includegraphics[width=\linewidth]{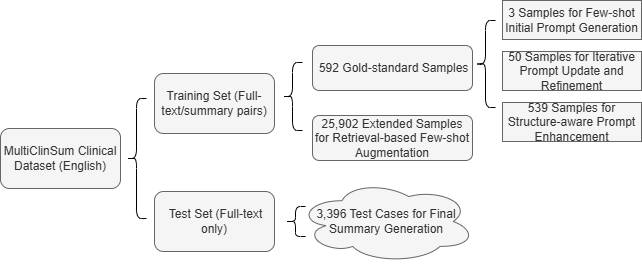}
  \caption{Data Partitioning and Usage in the MultiClinSUM Pipeline}
  \label{fig:Data Partitioning and Usage in the MultiClinSum Pipeline}
\end{figure}

For LLMs, we used GPT-4 to generate the initial prompt based on a meta-prompt,  
and GPT-4o for summarisation and reflection generation. We split the official MultiClinSUM data into several parts for training around 5 epochs, and selected a well-performing prompt for final test set inference, as described in Section \ref{sec: Methodology}. Figure \ref{fig:Data Partitioning and Usage in the MultiClinSum Pipeline} illustrates the data partitioning strategy and its corresponding usage in our pipeline. We completed summary generation for 3,396 English full-text cases. Other languages (Spanish, Portuguese, and French) will be explored in future work.

\subsection{Submission outcome of \textsc{MaLei} from the shared task}
\subsubsection{Quantitative Results}

\begin{figure}[htbp]
  \centering

  \begin{minipage}[t]{0.48\linewidth}
    \centering
    \includegraphics[width=\linewidth]{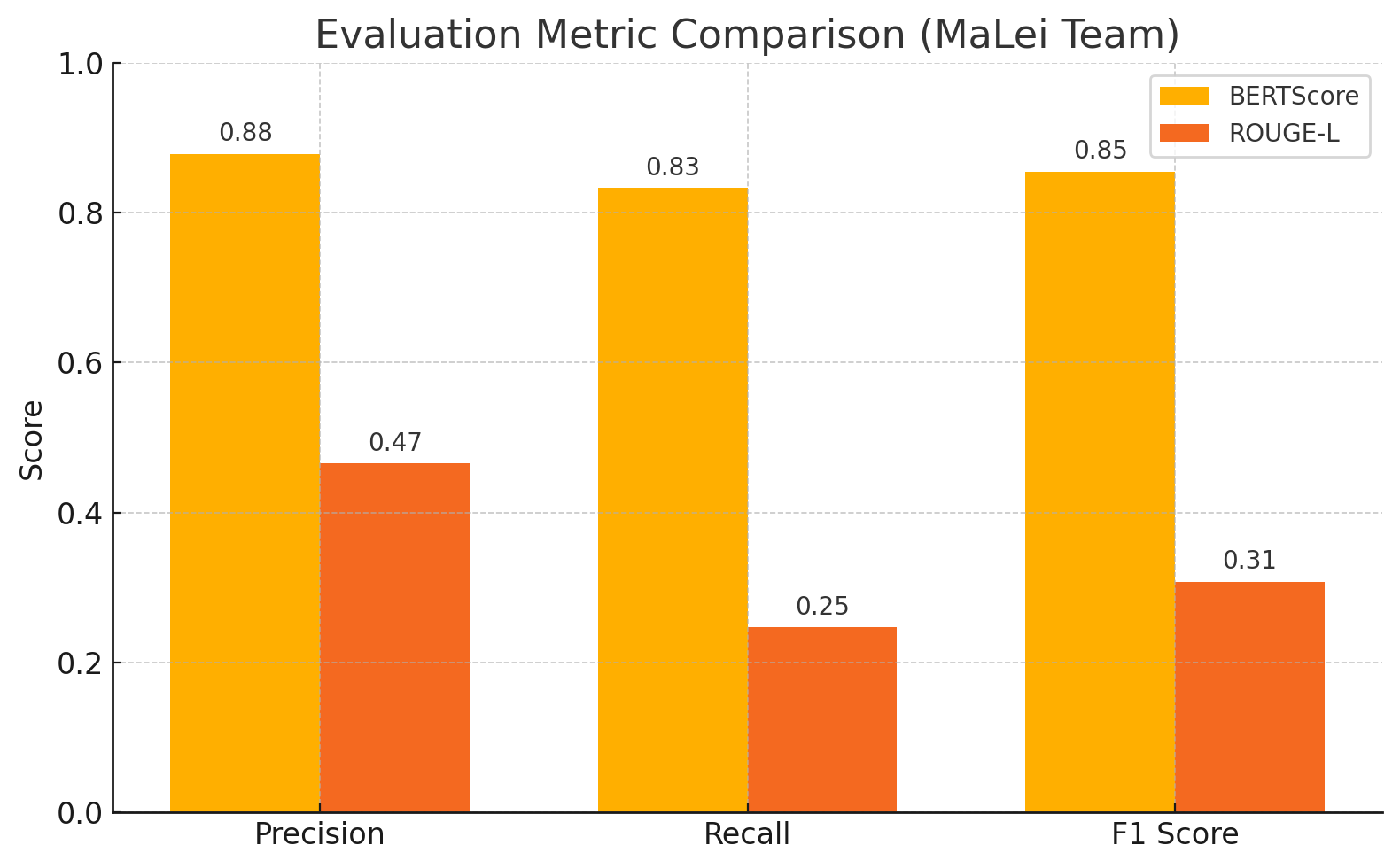}
    \captionsetup{width=0.95\linewidth, justification=centering}
    \caption{Evaluation Metric Comparison:\\ BERTScore vs ROUGE-L}
    \label{fig:metric_comparison}
  \end{minipage}
  \hfill
  \begin{minipage}[t]{0.48\linewidth}
    \centering
    \includegraphics[width=\linewidth]{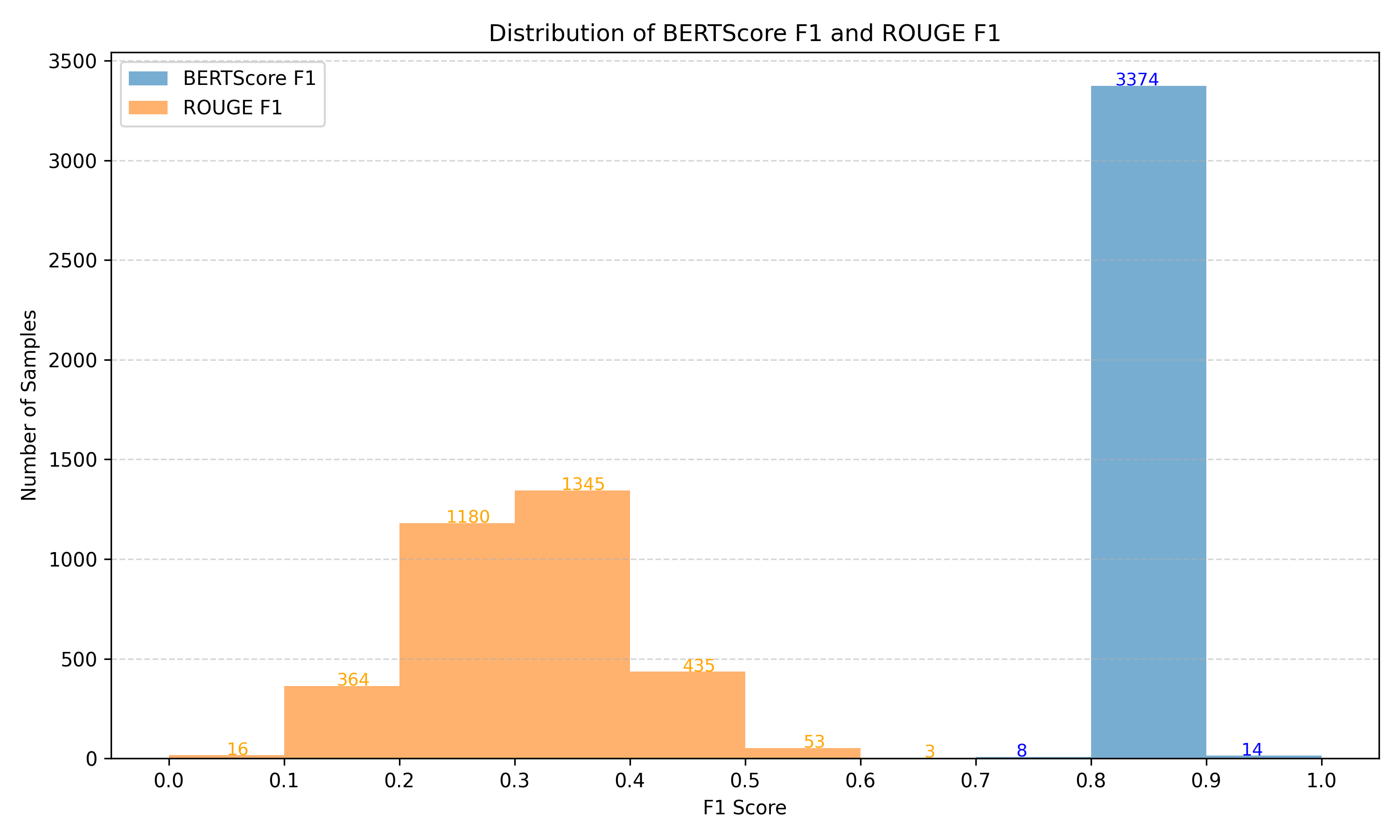}
    \captionsetup{width=0.95\linewidth, justification=centering}
    \caption{F1 Score Distribution: BERTScore vs ROUGE}
    \label{fig:f1_distribution}
  \end{minipage}

\end{figure}

For the MultiClinSUM2025 shared task that we attended \cite{BioASQ2025MultiClinSum}, the results of 3,396 submitted English test summaries are shown in Table \ref{tab:evaluation-metrics}, and their corresponding Grouped Bar Chart and Overlaid Histogram are shown in Figure \ref{fig:metric_comparison} and Figure \ref{fig:f1_distribution}.

\begin{table}[!ht]
\centering
\caption{\centering Evaluation metrics for BERTScore and ROUGE from MaLei team submissions.}
\begin{tabular}{lccc}
\hline
\textbf{Metric} & \textbf{Precision} & \textbf{Recall} & \textbf{F1 Score} \\
\hline
BERTScore & 0.8784 & 0.8325 & 0.8546 \\
ROUGE-L     & 0.4653 & 0.2468 & 0.3077 \\
\hline
\end{tabular}
\label{tab:evaluation-metrics}
\end{table}

At the test set level, as shown in Table \ref{tab:evaluation-metrics} and Figure \ref{fig:metric_comparison}, BERTScore is overall more than twice as high as ROUGE-L, reflecting a similar trend observed in the training set. This suggests that our system achieves strong semantic preservation while tending to paraphrase the original full text using different linguistic styles. Another notable pattern further supports this. Across both metrics, precision consistently exceeds recall. This indicates that the synthetic summaries are generally accurate in terms of what they include, but may lack completeness at a finer-grained level. We speculate that this may be because GPT-4o tends to generate more concise or compressed text. Additionally, the ROUGE-L recall is particularly low — falling below 0.25 — which implies that the model often uses more varied expressions instead of preserving the original key phrases, leading to reduced lexical overlap. Therefore, future work could focus on identifying and preserving fixed phrases and structural patterns in the summary generation process. In summary, this result aligns with what has been observed in autoregressive models: they tend to focus more on what to generate rather than on precise word-by-word reproduction.

At the instance level, as shown in Figure \ref{fig:f1_distribution}, nearly all samples (3374 out of 3396, $\approx 99.35\%$) have BERTScores concentrated in the 0.8–0.9 range, indicating a high degree of semantic consistency. In contrast, most ROUGE-L scores fall between 0.2 and 0.4. The distribution exhibits a clear left-skew and a long-tail pattern, suggesting that ISP may have inaccurately generalized the structure of the examples provided to the model, thereby misleading the LLM’s generation.
Notably, 380 samples (11.19\%) received ROUGE-L scores below 0.2, while none scored particularly low on BERTScore. This indicates that a subset of summaries may suffer from issues such as missing critical information, disorganized structure, or fragmented language.

\subsubsection{Qualitative Results}
\begin{figure}[!ht]
    \centering
    \includegraphics[width=\linewidth]{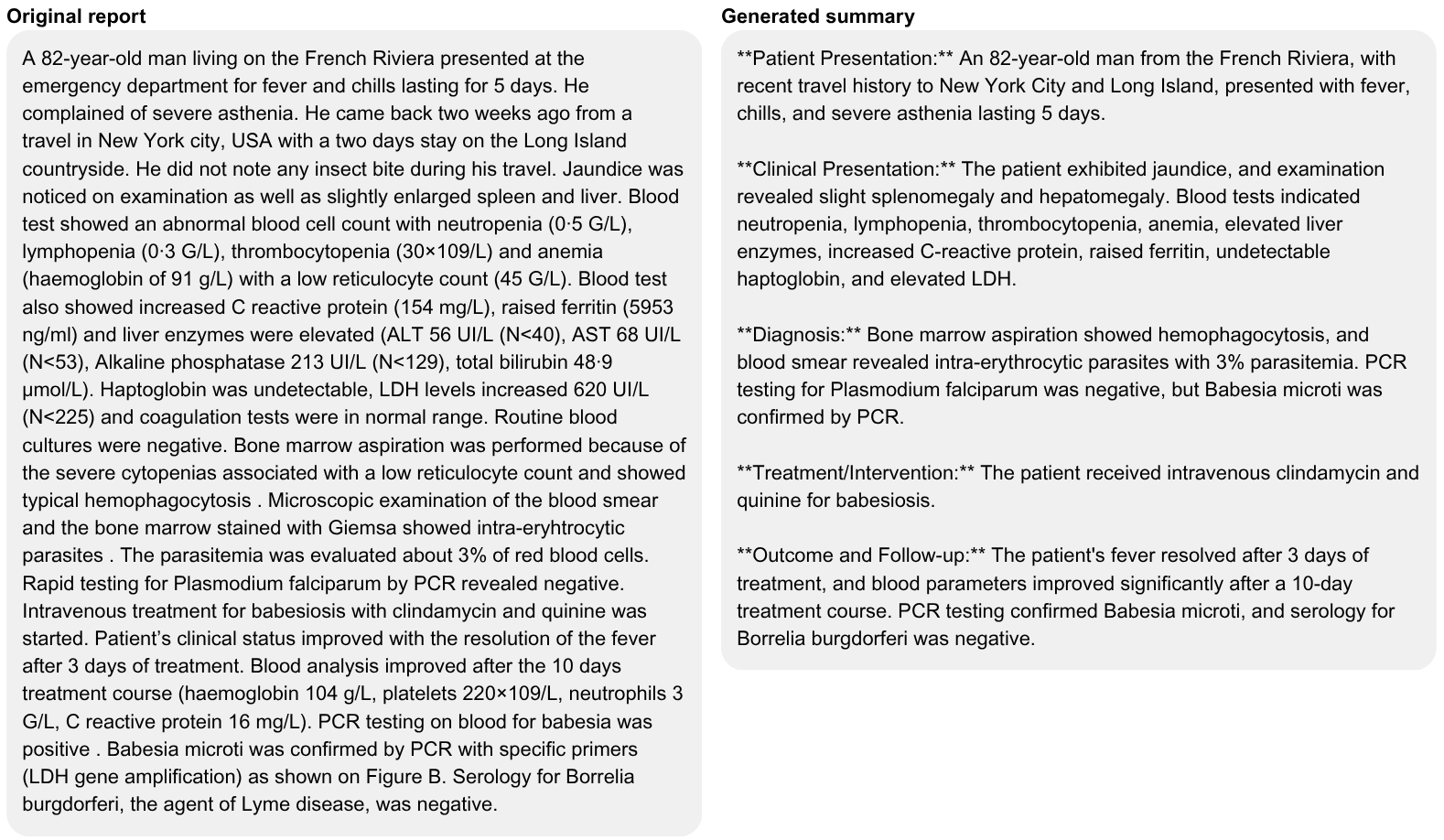}
    \caption{Example of an original report next to the generated summary using the ISP technique, in which the generated summary covers key aspects in the original report.}
    \label{fig:examplesum0}
\end{figure}
Generally, the summaries tend to cover key features in the clinical report and contain logical paraphrasing, which likely led to the high BERTscores, i.e., a high semantic overlap. Figure \ref{fig:examplesum0} shows how the generated summary covers key aspects about the patient, such as the age (82), gender (male), and the travel history that is relevant for his symptoms. The summary then describes the symptoms (jaundice, etc.), and the tests and results that have led to a particular diagnosis, followed by the treatment and outcome. Based on a qualitative analysis with a small sample, we report that most generated summaries appear to be well-structured according to key features present in human-written clinical reports and consistent with the original report, which confirms the high BERTscores. 

\begin{figure}[!ht]
    \centering
    \includegraphics[width=\linewidth]{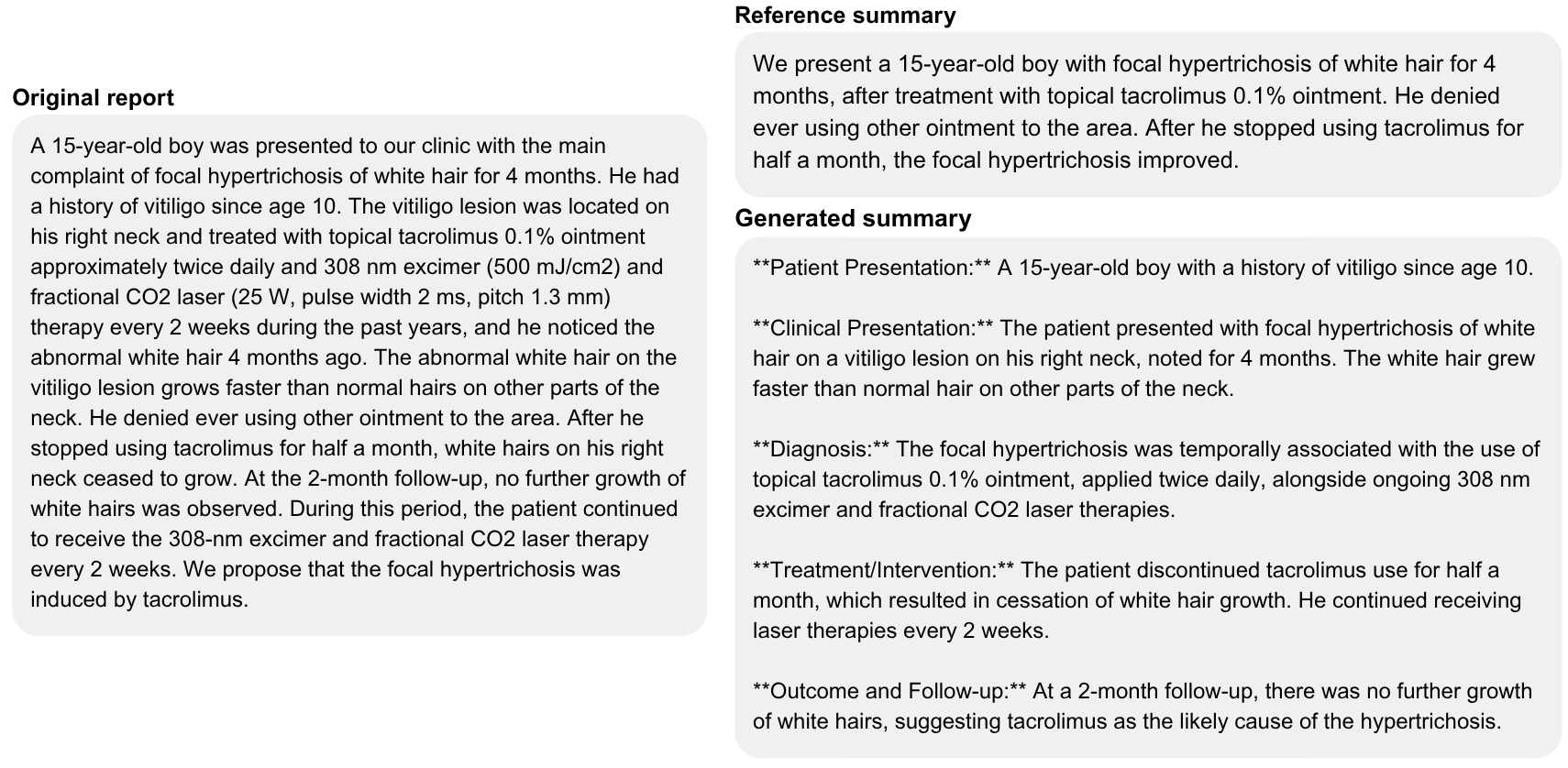}
    \caption{Example of output during prompting procedure in which the generated summary is much longer and detailed than the reference summary.}
    \label{fig:examplesum1}
\end{figure}

Comparing the generated summaries during the prompt updating phase and the reference summaries in the training set, we observe that the generated summaries tend to be longer than the gold-standard summaries. The generated summaries tend to be more detailed, including details about the specific tests done and the outcomes of these to reach a particular diagnosis. Figure \ref{fig:examplesum1} showcases an example in which the reference summary is much shorter than the LLM-generated summary. The generated summary contains more details regarding the different treatments that were done previously, while the reference summary focuses on the main complaint of the patient (``focal hypertrichosis of white hair") and the treatment that contributed to the patient's improvement (``discontinued tacrolimus use"). This suggests that the LLM struggles with discerning key events from details that might be redundant for domain experts, who may be able to infer what procedures were done from a short and dense summary. In addition, the generated summaries always introduce the full form of abbreviations, which is not always the case in gold-standard summaries. These differences might have led to a lower ROUGE-L score. 

\begin{figure}[!ht]
    \centering
    \includegraphics[width=\linewidth]{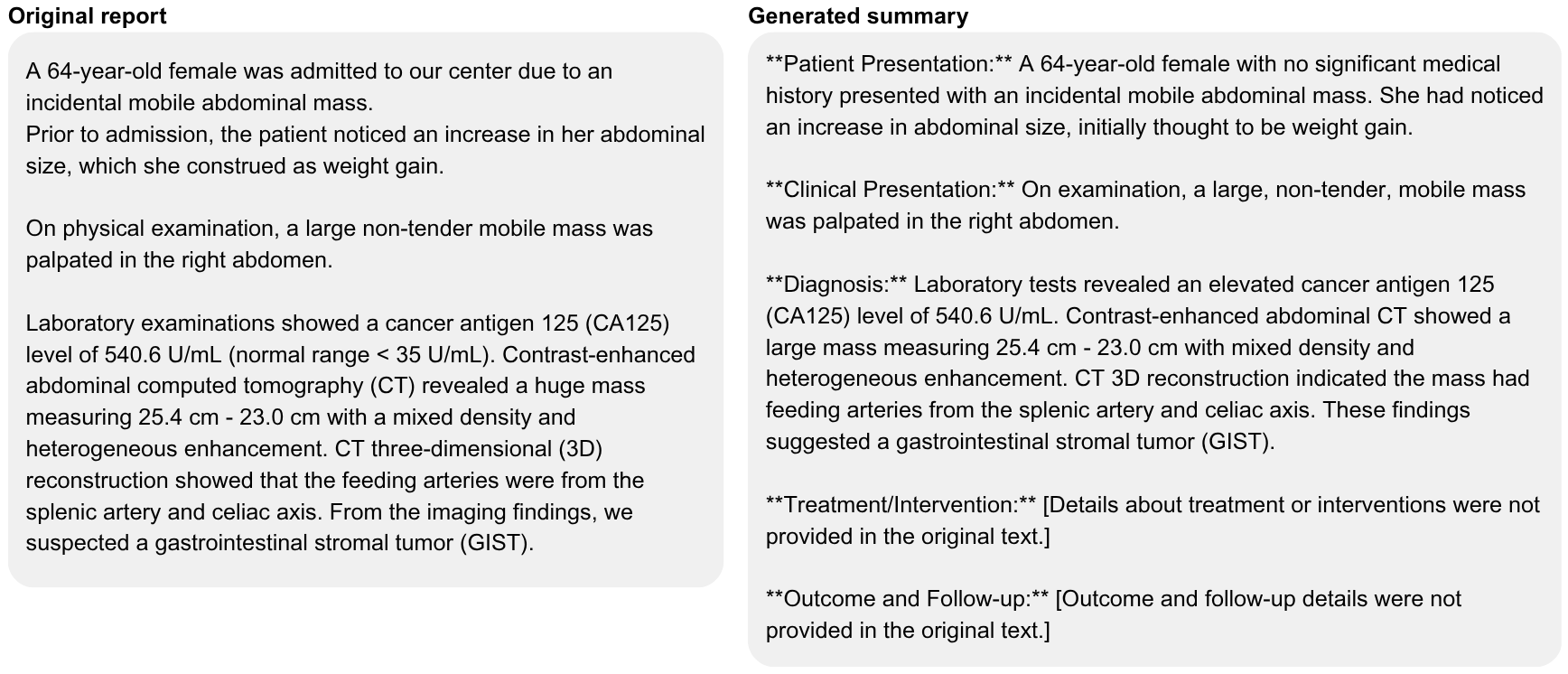}
    \caption{Example of an instance in the test set where the generated summary is longer than and very similar to the original report. }
    \label{fig:examplesum2}
\end{figure}

Furthermore, we found that our prompt design resulted in the LLM including section headers with every generated summary (see Figure \ref{fig:examplesum1} and Figure \ref{fig:examplesum2}). This led to the generated summaries strictly adhering to the structure as provided in the prompt, e.g., ``Patient presentation'', ``Diagnosis'', and ``Treatment''. This explicit structure was generally absent in gold-standard summaries in the training set. Moreover, the strict adherence to the provided structure might have led to an ordering of key events and entities that is different from the reference summary. While the structure is logical and follows key features of clinical reports and summaries, it might have negatively impacted the structural overlap between the generated summary and the reference summary. Therefore, in this case, the inclusion of the section headers and strict adherence to the structure provided in the prompt might have contributed to the overall low ROUGE-L score. 

Surprisingly, some generated summaries (12 texts in the test set, out of 3,396) are longer than the original texts (excluding the section headers). Examining the original clinical reports of this sample, it appears that the original reports are already quite brief and information-dense. The original texts of these generated summaries average around 135 words, which is much shorter than the average of 527 words in the entire test set. Close analysis of this sample reveals that the generated texts are mostly a repetition of the original text rather than a summary of key aspects and events. Figure \ref{fig:examplesum2} depicts how the generated summary copies phrases and sentences literally from the original report, only swapping a few verbs with close synonyms, e.g., ``revealed'' instead of ``showed''. 

The generated summaries also frequently include observations about missing information in the report, such as ``The case report does not provide specific details on the outcome or follow-up. Typically, such a patient would require close monitoring and treatment adjustments based on laboratory and clinical responses.'' Figure \ref{fig:examplesum2} exemplifies this. This likely negatively impacted the automatic evaluation scores, but might be useful for domain experts who, in this way, can gain insight into what key information is missing in the original report. 

\section{Discussion}
\subsection{Prompt-Driven Model Behavior}
With the iteratively updated prompt, the model exhibited the following behaviors:
\begin{itemize}
    \item \textbf{Limited ability to compress content, }especially in the \textit{Treatment} and \textit{Patient Presentation} sections. This may be due to the lack of content filtering—the model tends to treat all information equally, failing to prioritise critical conditions and key treatments.
    \item \textbf{Structure-guided prompting may induce hallucinations or additional content, }particularly due to the decoder-only architecture. For example, if the prompt asks the model to summarize the \textit{Outcome}, but the original text lacks such content, the model may fabricate information like "regular follow-up was scheduled." It may also lead to the model filling the gaps with statements such as "[Outcome and follow-up details are not provided in the original text."
    \item \textbf{Limited structural flexibility. }The model tends to follow the prompt-defined structure too rigidly, often generating key sentences by copying large portions of the original text with only minor adjustments. It is also prone to explicitly including section headings based on the focus points specified in the prompt, which may negatively affect ROUGE-L performance, particularly when the generated order differs from the gold reference.
\end{itemize}

Future work should focus on instructing the model to avoid redundant restatement, introducing counterfactual constraints to reduce hallucinated content, and developing more flexible structural control, such as preserving original abbreviations instead of expanding them, and softening the enforcement of fixed section headers.

\subsection{Reflections on Evaluation Metrics}
Although the original evaluation metric, ROUGE-L, effectively captures lexical overlap between the generated and gold summaries, it has notable limitations and may underestimate summary quality in certain cases. This is primarily because ROUGE-L is highly sensitive to variations in sentence structure and phrasing.

Furthermore, the manually annotated gold summaries are often highly compressed, frequently using abbreviations and omitting connective phrases. In contrast, the synthetic summaries tend to resemble patient-facing clinical reports, featuring more complete and explicit expressions. As a result, the two types of summaries may differ more in style than in substance. A low ROUGE-L score does not necessarily indicate poor summary quality, as the generated version may convey equivalent medical content in a different form.

Future work could incorporate metrics that account for structural coverage or introduce clinically grounded factual checks as complementary evaluation strategies.


\section{Conclusions and Future Work}
For this shared task on multilingual clinical document summarisation, we used perspective-aware iterative-self prompting (ISP) on LLMs via GPT4/4o, with the inspiration of the work by \cite{romero2025manchester}. During the model development, we designed the following perspectives for summarisation, including Patient Presentation, Clinical Presentation, Diagnosis, Treatment, and Outcome (Follow-up).
In conclusion, the perspective-aware iterative self-prompting (\textbf{PA-ISP}) on LLMs can help summarise lengthy clinical documents into short summaries while keeping the essence of the clinical knowledge, thus to help clinicians understand patients’ healthcare history more efficiently, and help the patients to understand their condition better.
The future work will also include lay/plain language adaptation into the summarisation so that patients with low health literacy level can better understand the clinical records, thus to improve the communications between patients and healthcare providers for better shared decision making.
Local LLMs will be explored and trained for better privacy preservation.
LLM Explainability and Reasoning are also our ongoing work.
In addition, we plan to consider other languages such as the Spanish data from the shared task, as well as comparing more diverse prompts.

\begin{acknowledgments}
We thank Ida Korfage, Associate Professor at Erasmus MC and co-PI from the 4D Picture Project, for the valuable comments and revision on the Conference Abstract version of this paper.
We thank the anonymous reviewers for their feedback on improving our article. 

\end{acknowledgments}

\section*{Declaration on Generative AI}
  
 %
 During the preparation of this work, the author(s) used GPT-4o in order to: Grammar and spelling check. 

\bibliography{sample-ceur}

\appendix

\section{\textsc{MaLei} Team Online Resources}

The sources for the \textsc{MaLei} Team at MultiClinSUM shared task 2025 will be available via 
\begin{itemize}
\item \href{https://github.com/Libo-Ren/MultiClinSum}{GitHub} \url{https://github.com/Libo-Ren/MultiClinSum},
\item
  \href{https://github.com/pabloRom2004/-PerAnsSumm-2025}{Our earlier PerAnSumm page from Manchester Bees} \url{https://github.com/pabloRom2004/-PerAnsSumm-2025}.
\end{itemize}

\section{Prompt}
\label{sec:prompt}
\subsection{Start Prompt}
\label{sec:prompt-example}
\begin{figure}
  \centering
  \includegraphics[width=0.9\linewidth]{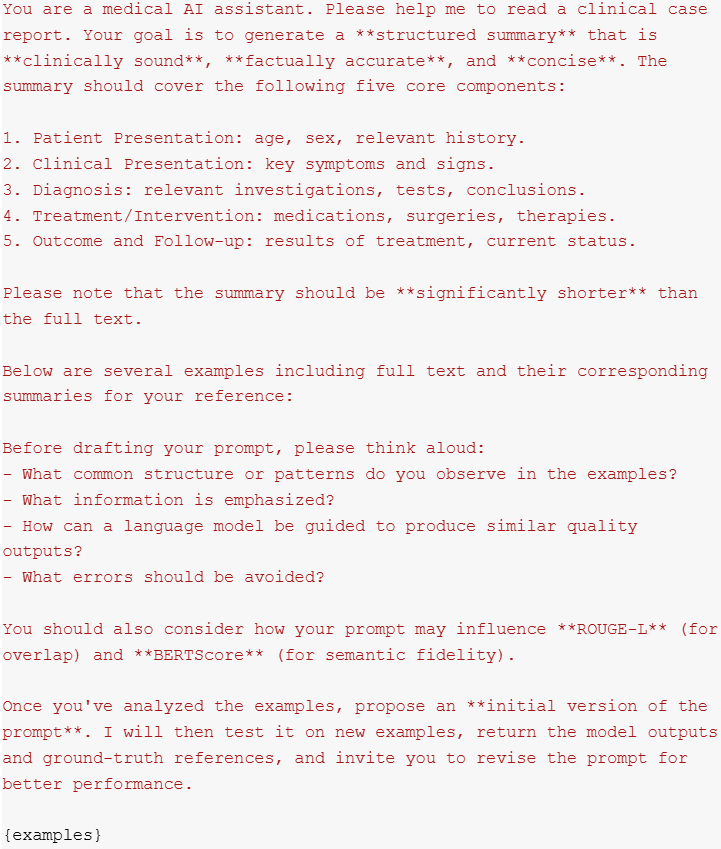}
  \caption{An example of starting prompt}
  \label{fig:starting prompt}
\end{figure}

An example of starting a prompt using ISP is shown in Figure \ref{fig:starting prompt}.

\subsection{Instruction for Prompt Refinement}
\label{sec: Instruction for Prompt Refinement}
\begin{figure}
  \centering
  \includegraphics[width=0.9\linewidth]{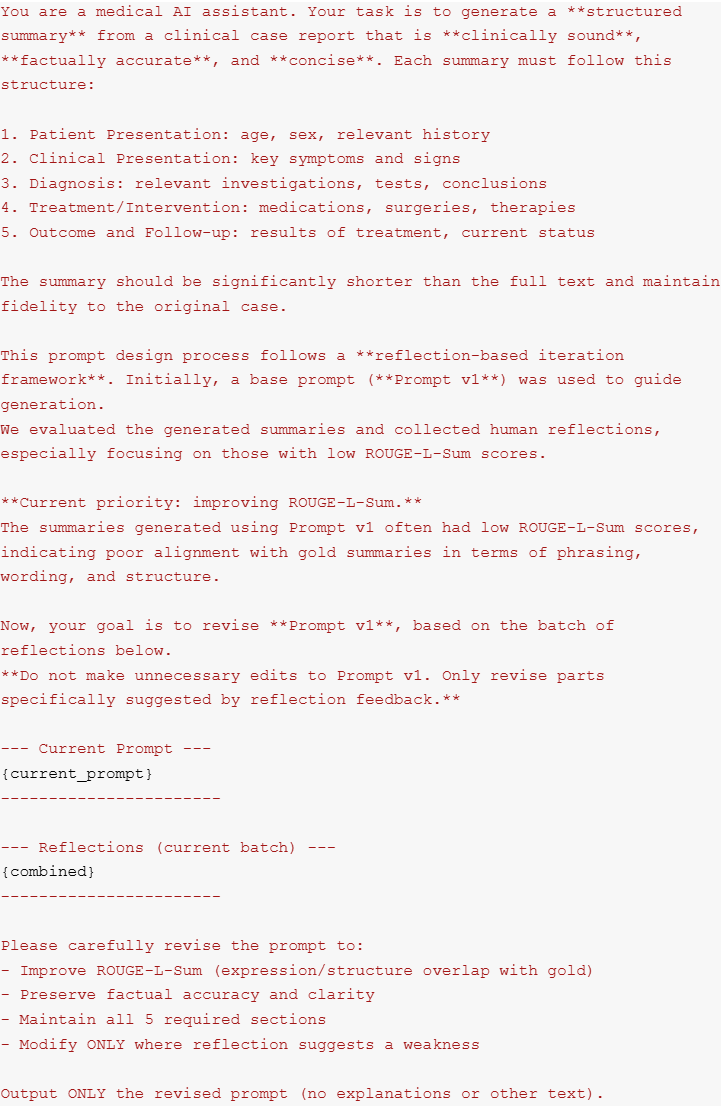}
  \caption{Instruction for Prompt Refinement}
  \label{fig:Instruction for Prompt Refinement}
\end{figure}

Meta-prompt used to instruct the LLM to revise the summary-generation prompt is shown in Figure \ref{fig:Instruction for Prompt Refinement}.

\end{document}